\documentclass[10pt,twocolumn,letterpaper]{article}

\usepackage{cvpr}
\usepackage{times}
\usepackage{epsfig}
\usepackage{graphicx}
\usepackage{amsmath}
\usepackage{amssymb}
\usepackage{amsopn}
\usepackage{multirow}
\usepackage{stfloats}
\usepackage{color}

\cvprfinalcopy

\DeclareMathOperator{\softmax}{\textrm{softmax}}
\renewcommand{\matrix}[1]{\mathbf{#1}}
\renewcommand{\vec}[1]{\mathbf{#1}}
\newcommand{\transpose}{\intercal}
\newcommand{\pheadB}[1] {\vspace{1mm}\noindent\textbf{#1}}
\newcommand{\tabincell}[2]{\begin{tabular}{@{}#1@{}}#2\end{tabular}}

\begin{document}

\title{Attention Clusters: Purely Attention Based\\ Local Feature Integration for Video Classification}

\author{Xiang Long\textsuperscript{1},
Chuang Gan\textsuperscript{1}\thanks{Corresponding author.},
Gerard de Melo\textsuperscript{2} ,
Jiajun Wu\textsuperscript{3} ,
Xiao Liu\textsuperscript{4} ,
Shilei Wen\textsuperscript{4} \\
\textsuperscript{1}{Tsinghua University} ,
\textsuperscript{2}{Rutgers University} ,
\textsuperscript{3}{Massachusetts Institute of Technology},
\textsuperscript{4}{Baidu IDL}
}

\maketitle

\begin{abstract}
    Recently, substantial research effort has focused on how to apply CNNs or RNNs to better extract temporal patterns from videos, so as to improve the accuracy of video classification. In this paper, however, we show that temporal information, especially longer-term patterns, may not be necessary to achieve competitive results on common video classification datasets. We investigate the potential of a purely attention based local feature integration. Accounting for the characteristics of such features in video classification, we propose a local feature integration framework based on attention clusters, and introduce a shifting operation to capture more diverse signals. We carefully analyze and compare the effect of different attention mechanisms, cluster sizes, and the use of the shifting operation, and also investigate the combination of attention clusters for multimodal integration. We demonstrate the effectiveness of our framework on three real-world video classification datasets. Our model achieves competitive results across all of these. In particular, on the large-scale Kinetics dataset, our framework obtains an excellent single model accuracy of 79.4\% in terms of the top-1 and 94.0\% in terms of the top-5 accuracy on the validation set. The attention clusters are the backbone of our winner solution at ActivityNet Kinetics Challenge 2017~\cite{bian2017revisiting}. Code and models will be released soon.
\end{abstract}

\section{Introduction}

    Video classification remains one of the prime challenges in computer vision as well as machine learning. It has received a substantial amount of attention in recent years, owing not least to its numerous potential use cases, such as video tagging, surveillance, autonomous driving, and stock footage search. Thanks to recent large datasets, e.g.\ YouTube-8M \cite{youtube8m} and Kinetics \cite{2si3d}, the recognition accuracy in video classification has advanced considerably, although the current state-of-the-art remains subpar in comparison with human performance.

    \begin{figure}[t!]
        \centering
        \includegraphics[width=0.45\textwidth]{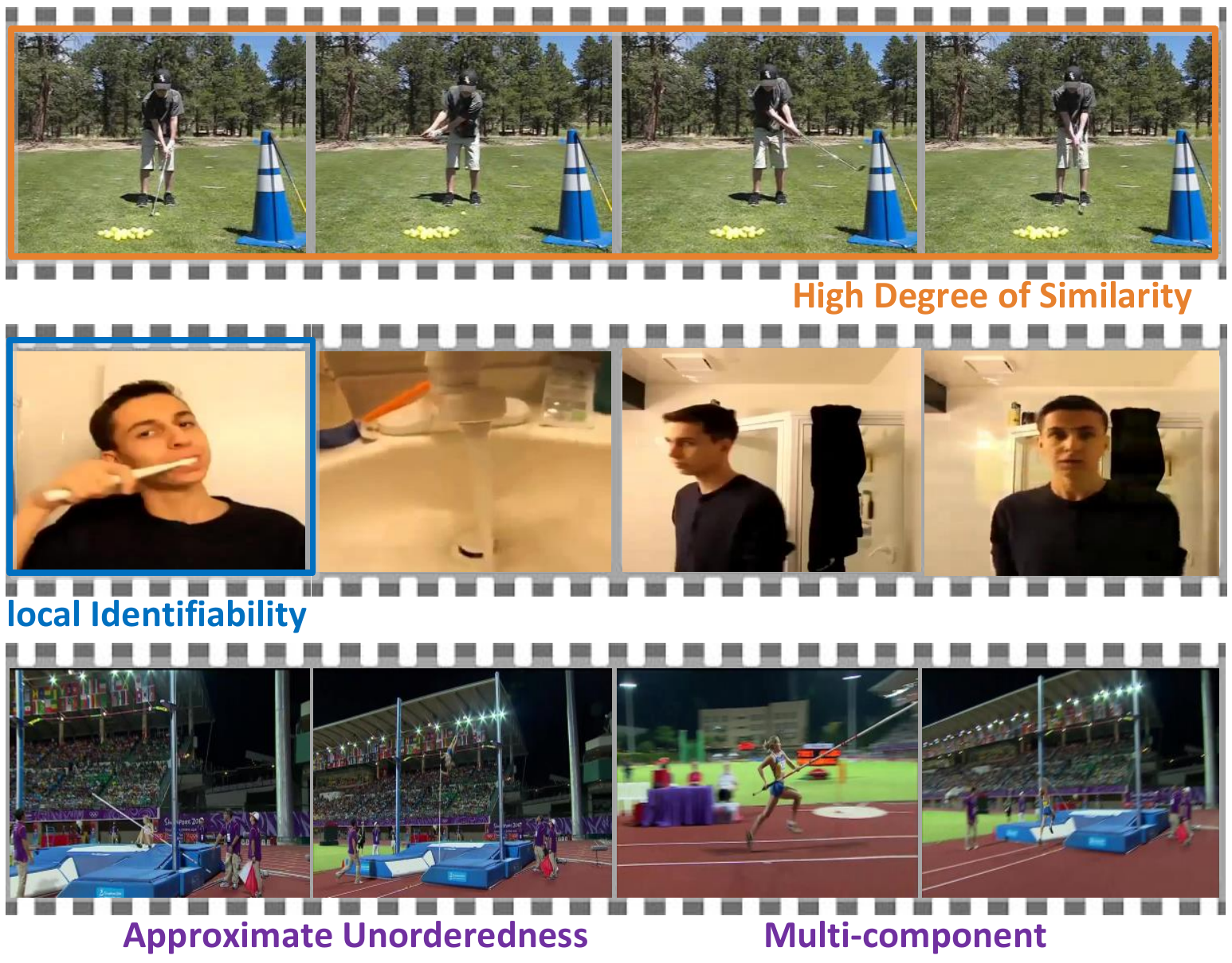}\\
        \caption{Example of RGB frames sampled from videos. The bottom ones are shown in a randomly permuted temporal order. We observe several important characteristics of local features of a video: high degree of similarity, local identifiability, approximate unorderedness, and multi-component inputs.}
        \label{fig:properties}
    \end{figure}
    Most of the existing effective video classification methods are based on convolutional neural networks (CNNs). CNNs have shown their powerful representation learning abilities in
    various image classification tasks. Convolutional and pooling layers together essentially act as potent feature extractors, which are able to mine local features from different regions of an image. Unsurprisingly, CNNs can also be used as a feature extractor for local feature extraction from videos, extracting a sequence of features for relevant video frames in accordance with their temporal order.

    Many existing methods use convolutional neural networks (CNNs) or recurrent neural networks (RNNs) based on such local feature sequences to capture the temporal interactions within a video. The latter are particularly often considered for their ability to capture longer-term temporal patterns by retaining pertinent state information across time. In this paper, however, we cast some doubt on whether temporal patterns, especially long-term ones, are truly indispensable for common video classification tasks. This is motivated by the following observations regarding the characteristics of such local features for video classification.

    First, local features within a video tend to have a \emph{high degree of similarity} across frames. In short video clips, the changes between RGB frames tend to be small. Especially during phases of slow movement, adjacent RGB frames exhibit substantial redundancy, sometimes differing in just miniscule ways. For instance, in the short example clip portraying \emph{golf chipping}
    in Figure \ref{fig:properties} (top), the frames are almost identical except for gradual changes in the position of the club.
    For classification, it may suffice to view these similar features holistically,
    packaged as a whole, while disregarding the particular details of their evolution over time.

    Second, the local features of a video often possess the property of \emph{local identifiability}. When people watch videos, they are frequently able to classify them based on just a few, occasionally even just a single frame. For example, in the video for \emph{brushing teeth} in Figure \ref{fig:properties} (middle), we can infer the class having observed just the initial frame. That is to say, even a tiny fraction of local features may single-handedly provide exhaustive classification information. For classification, the key is to spot the relevant local features of these most informative frames, without needing to ponder over their temporal patterns.

    Third, the local features of a video may be regarded as \emph{approximately unordered} in many classification settings. Of course, the input video itself is ordered, and a thorough semantic interpretation of the portrayed narrative requires some level of understanding of the temporal progression. For classification problems, however, we conjecture that the order may not be crucial. Even if the local features are permuted, a correct classification of the video may remain achievable. For example, in the \emph{pole vault} video in Figure \ref{fig:properties} (bottom), the frames have been reordered, showing first the landing, then the jump, and finally the run up. Yet, humans can still easily categorize it. Hence, the order of local features may not need to be preserved in video classification.

    Accounting for the above considerations, we investigate an approach that completely abandons temporal cues. Instead, we explore the potential of purely attention based local feature integration methods to generate a global representation. This is because attention mechanisms naturally possess the following properties. First of all, attention outputs are essentially weighted averages, which implies that repeated local features will automatically be aggregated to reduce their redundancy. Secondly, an attention model may assign higher weights to significant local features so as to focus on a small number of key signals, and their local identifiability determines to what extent the classification results on a small number of key frames can be taken as a class label for the entire video. Finally, the inputs to an attention model are naturally unordered sets of varying sizes, which fits the properties of the local features, and also facilitates a generalization ability to varying numbers of local features.

    We also observe a further important property, the \emph{multi-component} nature of local features of a video. Multiple cues in the signal may simultaneously make important contributions towards enabling the classification of a given video. For instance, in the \emph{pole vault} video (Figure \ref{fig:properties} bottom), the landing, jump, and run up all may yield useful information. Combining the signals across these different aspects ought to be better than focusing on just one of them.

    A single attention unit can be viewed as focusing on just one aspect of the video, hence discarding a considerable amount of information. It turns out that it is near-impossible to achieve our aims with a single attention unit. Thus, we propose using multiple attention units to construct an \emph{attention cluster} that constitutes a global representation of the video. Furthermore, we find that attention clusters resulting from a simple concatenation of the outputs of attention units only lead to weak gains, making them an inefficient choice. Instead, we propose a very simple and efficient procedure, the \emph{shifting operation}, which effectively increases the diversity between attention units, speeding up the training efficiency and improving the classification accuracy.

    In the following, we first review pertinent related work in Section \ref{sec:related-work}. Then, in Section \ref{sec:approach}, we present our proposed attention clusters approach with the shifting operation, as well as our overall architecture for video classification. In Section \ref{sec:flash-mnist}, in order to analyze the effect of various attention cluster approaches and visualize the inner workings of the attention mechanism, we propose Flash--MNIST as a new toy dataset, and conduct various comparative experiments on it. Finally, we show the results of using attention clusters on challenging real-world video classification datasets in Section \ref{sec:video}.

\section{Related Work} \label{sec:related-work}

    \subsection{Attention Mechanisms}
        Attention networks were originally proposed on the basis of the REINFORCE algorithm. In particular, Mnih et al.\ \cite{mnih2014recurrent} and Ba et al.\ \cite{ba2014multiple} proposed attention for object recognition with recurrent neural networks. These attention networks select regions by making hard binary choices, which may face difficulties in training.

        Soft attention mechanisms were proposed by using weighted averages instead of hard selections. Bahdanau et al.\ \cite{bahdanau2014neural} apply soft attention to machine translation with the aim of capturing soft alignments between source and target words. Sharma et al.\ \cite{sharma2015action} proposed a Soft-Attention LSTM model built on top of multi-layered RNNs to selectively focus on parts of the video frames and classify videos after taking a few glimpses. Li et al.\ proposed an end-to-end sequence learning model called VideoLSTM \cite{Li2016VideoLSTM}, which hardwires convolutions in the Soft-Attention LSTM. These soft attention models require the introduction of supplementary sources of information to guide the weighted averages, which incur a substantial computational cost while failing to yield sufficient improvements in classification tasks.

        To address the problem of attention on single sequences, many self-attentive models have been proposed for a variety of tasks, such as reading comprehension \cite{Cheng2016Long} and abstractive summarization \cite{2017arXiv170504304P}. Lin et al.\ \cite{Lin2017A} applied multiple attention units to learn task-independent sentence representations, relying on a penalization term to force each attention to attend to different parts. However, penalty functions forcing each weight vector to be different are too restrictive for video classification. Due to highly similar features between frames, many videos lack sufficient diversity, so this method fails to obtain good results. Our proposed attention clusters are another form of self-attentive architecture, which introduces a shifting operation to learn diversified attention units.

    \subsection{Video Classification}

        Since CNNs enjoy great success in image classification \cite{alexnet,inception,vgg16,resnet}, they have also been applied to video classification tasks. Karpathy et al.\ \cite{Karpathy2014Large} studied multiple fusion methods based on pooling local spatio-temporal features extracted by 2D CNNs from RGB frames.
        This can be viewed as a preliminary exploration of the idea of integrating local feature sets, although simple pooling methods do not yield significant gains.

        Many architectures have been proposed for modeling spatio-temporal information. The optical flow method \cite{optical-flow} captures temporally local information by considering the variation in the surrounding frames. Simonyan et al.\ \cite{Simonyan2014Two} devised a method that uses both RGB and stacked optical flow frames as appearance and motion signals, respectively. The accuracy is significantly boosted even by simply fusing probability scores, which indicates that optical flow can contribute useful short-term motion information.
        Gan et al.\ \cite{gan2015devnet} proposed a cross-frame max pooling approach to capture the dynamic temporal information.
        Feichtenhofer et al.\ \cite{Feichtenhofer2016Convolutional,Feichtenhofer2016Spatiotemporal} compared a number of ways of fusing CNNs both spatially and temporally and combined them with ResNets \cite{resnet} to extract better spatio-temporal information. C3D \cite{C3D} extends 2D CNNs by using 3D convolution kernels to capture spatio-temporal information. Varol et al.\ \cite{Varol2017Long} found that better results could be achieved by expanding the temporal length of inputs and using optical flows instead of RGB inputs for 3D CNNs. Carreira et al.\ \cite{2si3d} incorporated the Inception architecture \cite{inception} into 3D CNNs.

        To model long-term temporal interactions in video classification, recurrent neural networks (RNN), particularly long short-term memory (LSTM) \cite{lstm} have been applied in numerous papers. Ng et al.\ \cite{Ng2015Beyond} devised two-stream LSTMs. Donahue et al.\ \cite{donahue2015long} proposed an end-to-end architecture based on LSTMs. Srivastava et al.\ \cite{Srivastava2015Unsupervised} attempted to improve the representation ability of LSTMs by first pre-training them in an unsupervised manner to reconstruct the input. Gan et al.\ \cite{gan2016you,gan2016webly} investigated to train the LSTMs with the Web noisy data.  However, the accuracy on video classification with these RNN-based methods has been unsatisfactory, which may indicate that long-term temporal interactions are not crucial for video classification. Our proposed method explores the potential of local feature integration without any recourse to long-term order information.

\section{Approach} \label{sec:approach}

    We now describe our approach of using attention clusters with a shifting operation, and show how to apply it to the task of video classification. We broadly consider three major parts: local feature extraction, local feature integration, and global feature classification. Each of these is addressed by suitable neural networks. Among them, the local feature extraction uses existing CNNs, and the global feature classification invokes fully connected and softmax layers. The main contribution lies in the local feature integration step, that is, our investigation of how to generate global representations given a set of local features.

    \subsection{Local Feature Set}

    In neural network tasks, we often obtain local features of a video, since CNNs can naturally be used as a feature extractor.
    Directly operating on a set of local features may be preferable to forcing their description into fixed-length feature vectors, especially for videos of varying lengths.

    The local feature set is defined as a set of unordered local features corresponding to different parts of the same video.
    Here, for convenience, we use a $L \times M$ matrix $\matrix{X}$ to represent a set containing $L$ local features, each row of which is a separate local feature vector $\vec{x}_i$:
    \begin{equation}
    	\matrix{X} = (\vec{x}_1, \vec{x}_2,..., \vec{x}_L)
    \end{equation}
    Note that, in fact, the set of local features is unordered, and hence permuting the rows of the matrix should not affect the results. Also, the number of local features $L$ can vary across different objects. The challenge we seek to address at this point is how to generate fixed-length global vectors $\vec{g}$ to classify objects based on pertinent information from the local feature sets as given by $\matrix{X}$.

    \subsection{Attention}

    We rely on an attention mechanism to obtain such global features. In classification settings, the attention is static, and the input contains only the local feature vector set itself. Its responsibility is to first analyze the importance of each local feature and then to bestow the global feature with as much useful information as possible, while ignoring irrelevant signals and noise. Such attention outputs can essentially be regarded as weighted averages on a vector set:
    \begin{equation} \label{eq:att}
    	\vec{v} = \vec{a} \matrix{X},
    \end{equation}
    where $\vec{a}$ is a weight vector of dimension $L$, which is determined by a weighting function.

    The choice of weighting function is the most crucial design decision to be made. Its input is the local feature set $\matrix{X}$, while its output is the weight vector $\vec{a}$, whose $\ell_1$ norm is 1. Each dimension of the weight vector corresponds to a local feature.

    There are many methods to compute the weights of local features. For instance, global averages can be considered as a degenerate form of attention, and the corresponding weighting function can be expressed as:
    \begin{equation} \label{eq:aver}
    	\vec{a} = \frac{1}{L} \vec{1},
    \end{equation}
    where $\mathbf{1}$ is a vector of dimensionality $L$ with all elements equal to 1.
    For a more malleable attention weighting function, we can use a single fully-connected layer that has only one cell (FC1), such as:
    \begin{equation} \label{eq:fc1}
    	\vec{a} = \softmax (\vec{w} \matrix{X}^\transpose + \vec{b}),
    \end{equation}
    where $\vec{w}$ and $\vec{b}$ are parameter vectors of dimensionality $M$ and $L$, respectively.
    Similarly, we may use two successive fully-connected layers of size $H$ and one hidden cell (FC2):
    \begin{equation} \label{eq:fc2}
        \vec{a} = \softmax \left(\vec{w}_2 \tanh(\matrix{W}_1 \matrix{X}^\transpose + \vec{b}_1)+\vec{b}_2\right),
    \end{equation}
    where $\matrix{W}_1$ is a parameter matrix of dimensionality $H \times M$, $\vec{b}_1$, $\vec{w}_2$ are parameter vectors of dimensionality $H$, and $\vec{b}_2$ are parameter vectors of size $L$.

    In the experiments in Section \ref{sec:flash-mnist}, we compare the effects of these different weighting functions.

    \subsection{Attention Clusters}
    The output of one such attention unit typically focuses on a specific part of the video, e.g.\ a particular set of related frames or similar sounds. Normally, a single attention unit can only be expected to reflect one aspect of the video. However, there can be multiple pertinent parts in a video that together describe the overall event portrayed in the entire video. Therefore, to be able to represent multiple components, we need multiple attention units that focus on different parts of local features. We refer to a group of attention units that operate on the same input but have independent parameters as an \emph{attention cluster}. The size $N$ of an attention cluster is defined by the number of independent attention units in it. The global feature $\vec{g}$ resulting from an attention cluster is a vector of dimensionality $NM$ which can be obtained by concatenating the outputs of all involved attention units:
    \begin{equation}
    	\vec{g} = [\vec{v}_1, \vec{v}_2, ..., \vec{v}_N],
    \end{equation}
    where $\vec{v}_k$ is output of $k$-th attention unit.

    \subsection{Shifting Operation}
    Although we expect attention clusters to be able to focus on different components, through experiments, which we describe in Section \ref{sec:shift}, we found that simply concatenating the outputs of attention units yields unsatisfactory results, since they tend to focus on similar signals. In order to address this problem, we propose a \emph{shifting operation}, which is added onto each attention unit.
    This is achieved by adapting Eq.~\ref{eq:att} as follows:
    \begin{equation}
    	\vec{v} = \frac{\alpha \cdot \vec{a} \matrix{X} + \beta} { \sqrt{N} \left \lVert \alpha \cdot \vec{a} \matrix{X} + \beta \right \rVert_2}
    \end{equation}
    where $\alpha$ and $\beta$ are learnable scalars, which act as a linear transformation in the feature space. After the linear transformation, we $\ell_2$-normalize each attention unit separately. The factor $1/\sqrt{N}$ finally acts as a global $\ell_2$-normalization on the cluster. Combining the linear transformation and normalization, the shifting operation shifts the weighted sum in the feature space and at the same time ensures scale-invariance. The shift operation efficiently enables different attention units to flexibly diverge from each other and have different distributions, and the scale-invariance facilitates the optimization of the entire network.

    \begin{figure}[t]
        \centering
        \includegraphics[width=0.46\textwidth]{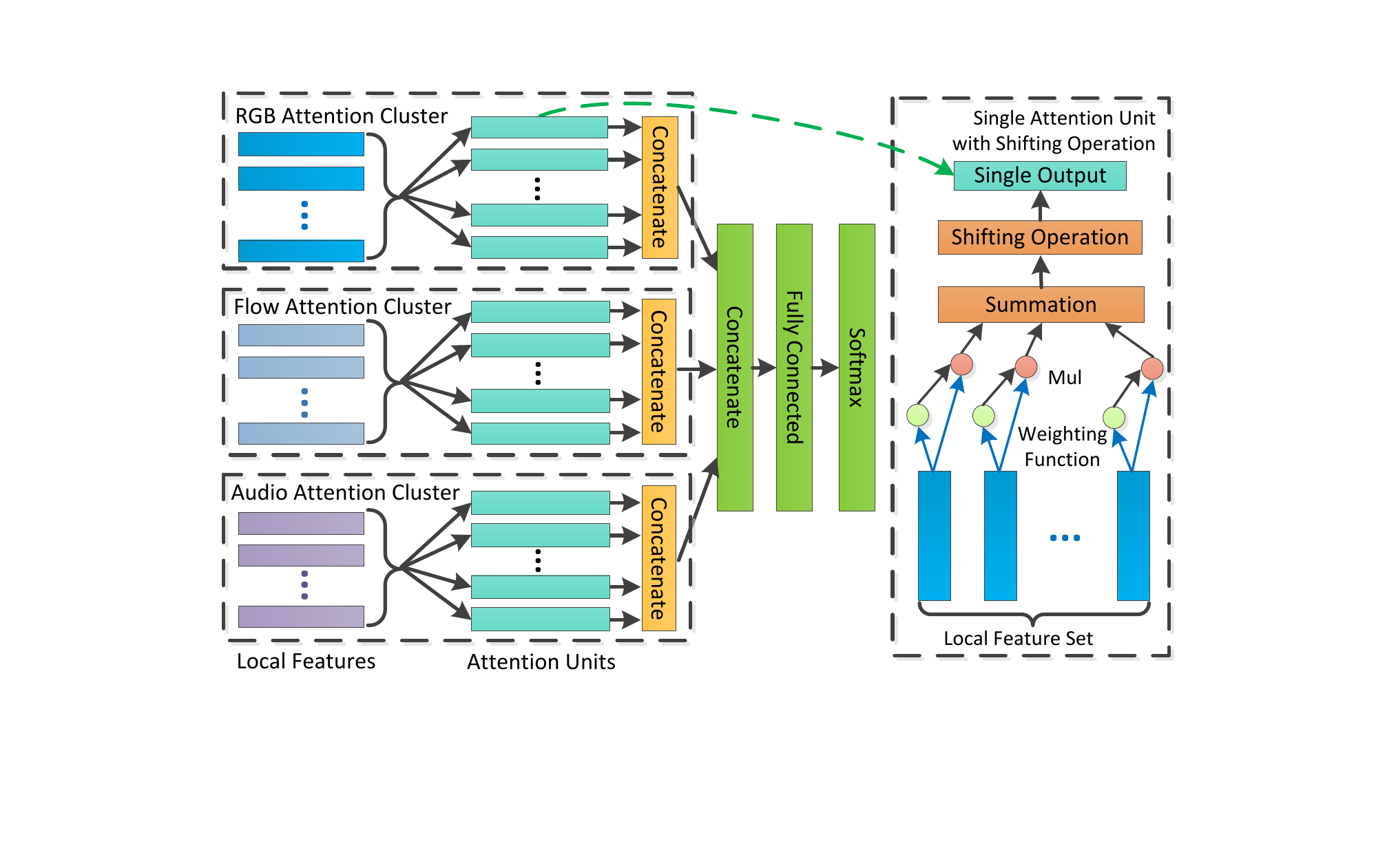}\\
        \caption{Multimodal Attention Clusters with Shifting Operation: The overall architecture for video classification. Separate attention clusters are applied for different feature sets and then the outputs are concatenated for classification.}
        \label{fig:satt}
    \end{figure}

    \subsection{Overall Architecture for Video Classification}
    In order to collect multimodal information from videos, we extract a variety of different local feature sets, such as appearance (RGB), motion (flow), and audio signals.
    However, it is unrealistic to process all feature sets simultaneously within the same attention cluster, because features of different modalities have different distributions, dimensionalities, and scales.
    Instead, we propose multimodal attention clusters with the shifting operation to train attention clusters for different modalities simultaneously.
    The layout of the proposed overall architecture is illustrated in Figure \ref{fig:satt}.

    First, we extract multiple feature sets from the video. For each feature set, we apply independent attention clusters with shifting operation to obtain a modality-specific representation vector. Next, the output of all attention clusters are concatenated to form a global representation vector of the video. Finally, the global representation vector is used for classification through a fully-connected layer.

\section{Analysis and Visualization} \label{sec:flash-mnist}
    Because real-world video classification datasets conflate many different forms of variation, they are not easy to analyze and visualize directly. We first propose a new toy video classification dataset, Flash--MNIST, which we synthetically generate from the MNIST handwritten digit dataset. The Flash--MNIST dataset has fewer irrelevant factors of influence and requires only modest amount of computation. Nevertheless, its local feature set has the same properties as for real video classification, which is convenient for analysis and visualization. This allows us to observe the behavior of the model under simplified conditions and achieve a deeper understanding of how the model works.

    \subsection{Flash--MNIST Dataset}

    \begin{figure}[t]
        \centering
        \includegraphics[width=0.44\textwidth]{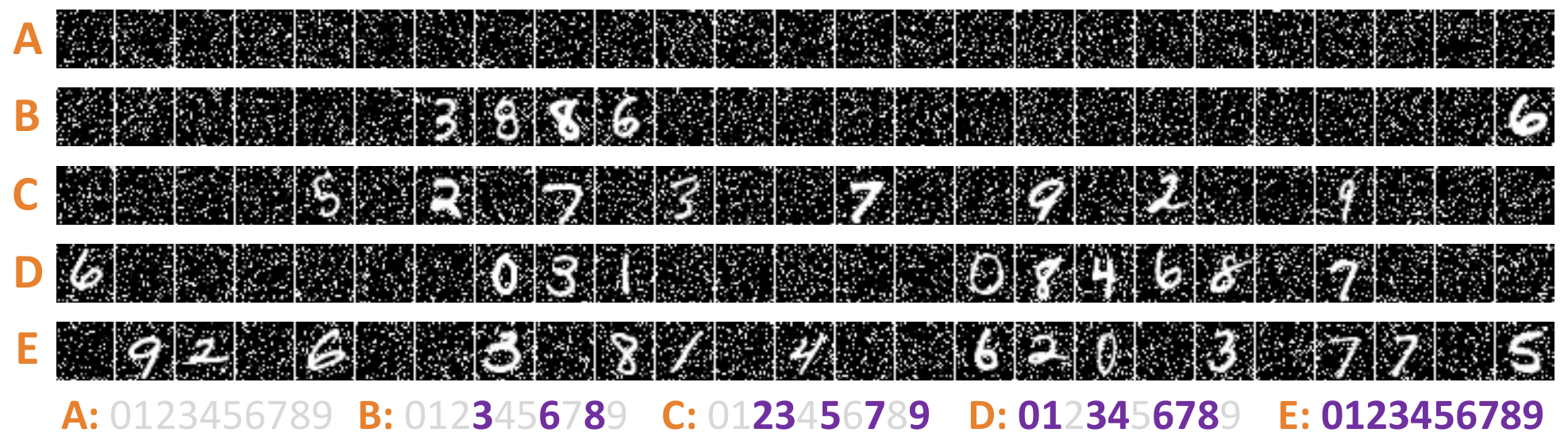}\\
        \caption{Frames of 5 videos in Flash-MNIST and the corresponding label. The \textcolor[rgb]{0.44,0.188,0.627}{\textbf{purple bold digits}} in label are digits flashing in the video.}
        \label{fig:mnist-example}
    \end{figure}

    In the well-known MNIST  dataset, the goal is classify $28 \times 28$ pixel images
    of handwritten digits into 10 classes for the respective digits.
    Flash--MNIST extends this image classification task to video. The videos in Flash--MNIST consist of 25 frames with noisy backgrounds on which various MNIST digits briefly flash up. The goal is to identify the specific set of digits that appear in the video, which entails choosing from a total of $2^{10}=1024$ categories. Figure \ref{fig:mnist-example} shows 5 samples and their corresponding labels. Specifically, in order to generate training samples, we first randomly generate 25 different $28 \times 28$ noise frames, sample a possible set of digits, and then randomly select corresponding digit images from the MNIST training set and overlay them on the random frames. For test set samples, a similar process is used, except that images are selected from the MNIST test data. We randomly generate 102,400 samples for training, and 10,240 samples for testing.

    We pretrain CNNs on MNIST with noisy backgrounds to extract local features. The CNNs consist of 2 successive convolutional layers with $5 \times 5$ kernels, 10/20 filters followed by \emph{relu} activations and max-pooling with stride 2, and one fully connected layer with 50 hidden units. Through the CNNs, we can obtain 25 local features with a dimensionality of 50, each local feature corresponding to a frame in the video. These 25 local features consist of the local feature set, and we apply a variety of attention cluster alternatives, as described in Section \ref{sec:approach}, to induce a global representation. Finally, this is passed through one fully connected layer for classification. The accuracy scores for different settings for the attention clusters is given in Table \ref{tab:mnist-result}. We describe and analyze the results in following. For more details of the dataset generation and network training, please refer to the Section \ref{sec:detail-mnist}.

    \begin{table}[t]
        \centering
        \resizebox{0.35\textwidth}{!}{
        \begin{tabular}{c|c|c|c|c|c}
        \hline
        \multirow{2}{*}{ \ $N$ \ } & \multirow{2}{*}{Average}
        &\multicolumn{2}{c|}{Without Shifting} & \multicolumn{2}{c}{With Shifting}\\
        \cline{3-6}
        & & \ \ FC1 \ \ & \ \ FC2 \ \ & \ \ FC1  \ \ &  \ \ FC2 \ \ \\
        \hline
        1  & 0.2  & 0.2  & 0.2  & 51.2  & 53.7 \\
        2  & 0.4  & 0.4  & 0.5  & 64.8  & 66.8 \\
        4  & 0.6  & 2.2  & 2.3  & 75.9  & 76.8 \\
        8  & 0.9  & 31.7  & 22.5  & 80.6  & 83.1 \\
        16  & 1.2  & 82.3  & 82.0  & 86.9  & 84.9 \\
        32  & 2.4  & \textbf{83.3}  & \textbf{83.2}  & \textbf{87.1}  & 85.6 \\
        64  & 5.0  & 83.2  & 82.4 & \textbf{87.1}  & 85.6 \\
        128  & \textbf{8.9}  & 81.8  & 80.9 & \textbf{87.1}  & \textbf{85.7} \\
        \hline
        \end{tabular}
        }
        \caption{Accuracy (\%) on Flash-MNIST to show the effect of different weighting functions
        , various cluster sizes $N$, and aggregation with or without the shifting operation.}
        \label{tab:mnist-result}
    \end{table}

    \subsection{Effect of Weighting Function}
    First of all, we analyze the effect of the choice of weighting function. We consider Average as described in Eq.~\ref{eq:aver}, and two different attention weighting functions,
    FC1 as described in Eq.~\ref{eq:fc1} and FC2 with 10 hidden units as described in Eq.~\ref{eq:fc2}. As shown in Table \ref{tab:mnist-result}, we observe a significant gap between the results of using Average and the other attention weighting functions, which means that attention can play an effective role in this situation to focus on the parts that merit consideration.
    We also observe that FC2 fares slightly better than FC1 when the cluster size is small, but FC2 performs worse than FC1 when the cluster size is large. This, we speculate, may stem from the expressive power of attention clusters saturating as the size increases, even if the form of attention itself is simple enough. Considering that FC2 contains more parameters and requires more computation, but is unable to yield any benefits, we rely on the FC1 weighting function as the default in all subsequent experiments.

    \begin{figure}[t]
        \centering
        \includegraphics[width=0.45\textwidth]{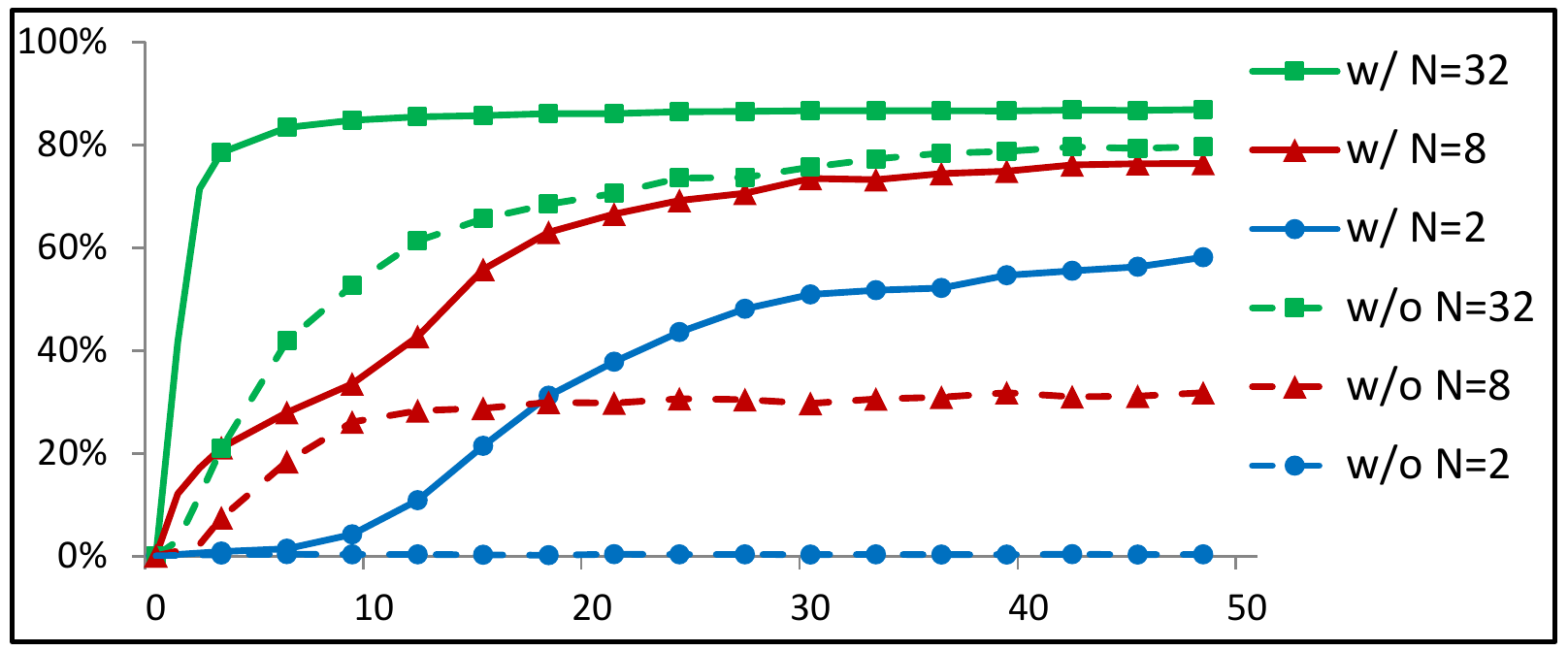}\\
        \caption{The accuracy on Flash--MNIST in each epoch, learned with different cluster sizes, with (w/) or without (w/o) the shifting operation.}
        \label{fig:mnist-converge}
    \end{figure}

    \subsection{Effect of Attention Cluster Size}

    Next, we consider the effect of different cluster sizes $N$.
    Because a video may include a variety of digits, these should not be attended to by the same unit. Consider an ideal situation, in which 10 attention units each pay attention to whether a specific digit occurs. This is obviously more reasonable than using a single attention function. Of course, during training, we lack control over which attention unit learns information about which digit. Still, we may hope that using multiple attention units may have the potential to learn more beneficial information.

    In order to verify our idea, for fairness of comparison, we ensure that the numbers of network parameters are completely identical, except for parameters contained in the weighting functions, by also replicating the output vectors for the Average method $N$ times.
    As shown in Table \ref{tab:mnist-result} and Figure \ref{fig:mnist-converge},
    we find that with an increase in the cluster size $N$, the classification results increase significantly when the size is small and subsequently almost remain unchanged until reaching a certain level. Furthermore, the gap between using attention and Average becomes larger as $N$ increases, indicating that this improvement is not due to an increase in the number of network parameters, but that the model genuinely pays attention to different aspects of local features.

    Besides, the convergence speed also increases for increasing cluster sizes. Although a larger cluster size requires more computation, a smaller overall training time is required for reasonably large cluster sizes, since the computation of the attention and shifting operations is very efficient.

\begin{figure}[t]
        \centering
        \includegraphics[width=0.44\textwidth]{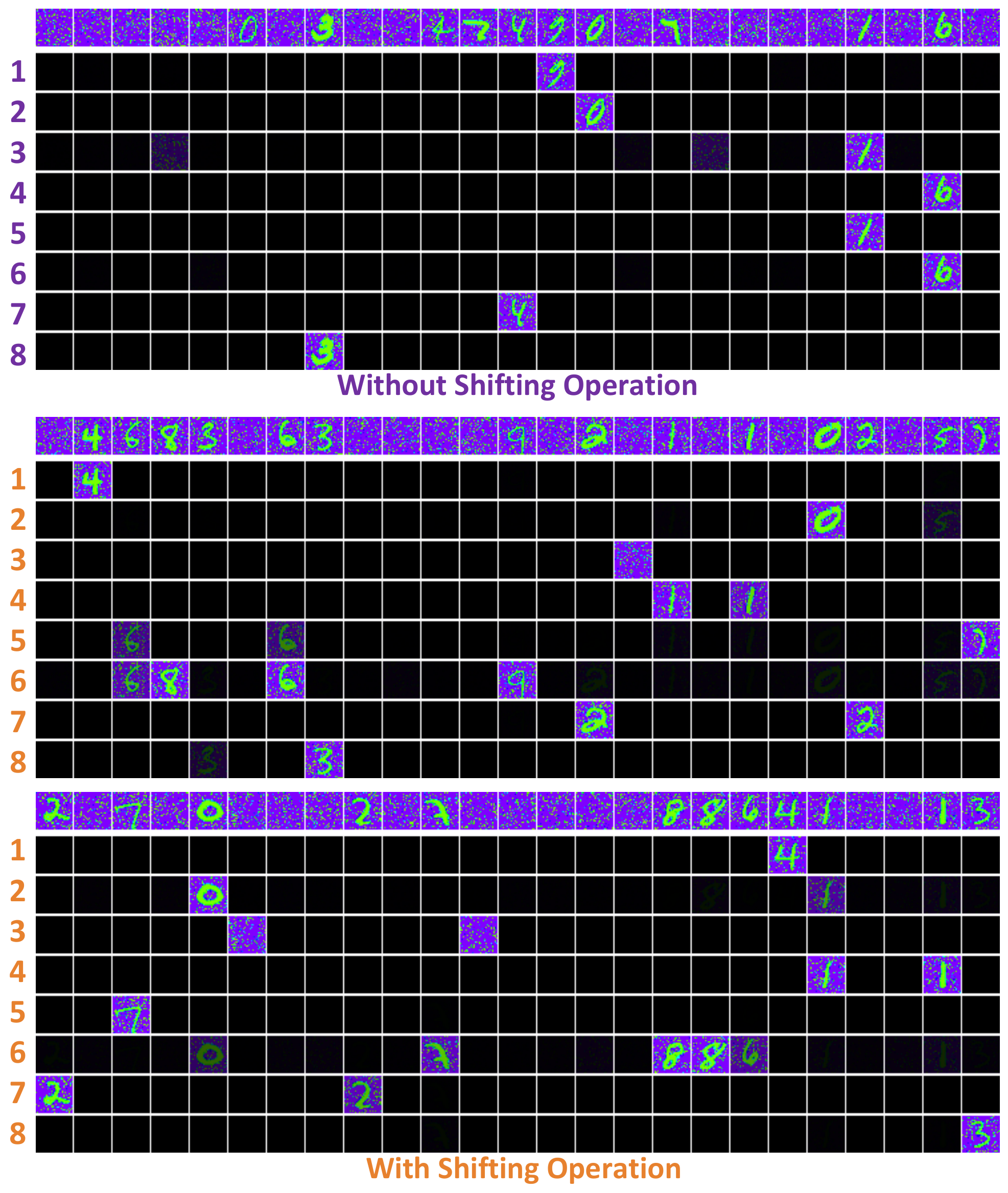}\\
        \caption{Visualization of attention weight maps of 8-unit attention clusters with shifting operation (middle, bottom), and without (top). We show frames in HSV space, in which larger values indicate a larger weight. The first row for each sample provides the video frames, and the respective following 8 rows correspond to the 8 weights of 8 attention units.}
        \label{fig:mnist-map}
    \end{figure}

    We visualize the attention weight maps for attention clusters with 8 units in Figure \ref{fig:mnist-map}. We observe that each attention unit learns about different kinds of information. For instance, with shifting operation (middle, bottom), the first attention unit learns to attend to the digit 4, and the fifth attention unit learns to attend to digits 6 and 7.

    \subsection{Effect of Shifting Operation} \label{sec:shift}

    Finally, we consider the effect of the shifting operation. As shown in Table \ref{tab:mnist-result}, we find that applying attention clusters with shifting yields substantial improvements, as the accuracy increases from 83.3\% to 87.1\%. Inspecting the attention weight maps in Figure \ref{fig:mnist-map}, we find that the attention weights diverge entirely when using shifting, while the third and fourth attention weights match the fifth and sixth when not using the shifting operation. This indicates that the shifting operation can help us learn more diversified information for better generalization, and, ultimately, a higher accuracy. Simultaneously, the shifting operation can also help the model converge more rapidly. As plotted in Figure \ref{fig:mnist-converge}, at the same cluster size, the approach with shifting converges much more rapidly than when forgoing the shifting operation.

\section{Experiment on Real Video Classification} \label{sec:video}

    In this section, we proceed to evaluate and compare our proposed methods on real-world video classification tasks.

    \subsection{Datasets}
        Specifically, we evaluate our methods on three popular trimmed video classification datasets.

        \pheadB{UCF101} \cite{ucf101} contains 13,320 web video clips with heterogeneous forms of camera motion and illumination. Each clip contains about 180 frames and is labeled with one of 101 action classes, ranging from daily life activities to unusual sports.

        \pheadB{HMDB51} \cite{Kuehne2011HMDB} consists of 6,766 video clips from movies and web videos. Each clip is labeled with one of 51 action categories.
        For UCF101 and HMDB51, we report the average accuracy over three training/testing splits, following the original evaluation scheme.

        \pheadB{Kinetics} \cite{2si3d} is a large-scale trimmed video dataset with more than 300K video clips in total, in which 246,535 serve as training data, and 19,907 for validation. Each video clip is taken from a different YouTube video and lasts around 10s. The clips are labeled using a set of 400 human action classes. The annotations for the test split have not yet been released, and we hence report experimental results on the validation split. Since Kinetics is larger-scale and has more categories, it is convenient for stable and reliable experimental analysis. We mainly perform comparative experiments on Kinetics.

    \begin{table}[t]
        \centering
        \resizebox{0.32\textwidth}{!}{
        \begin{tabular}{c|c|c|c|c|c|c}
        \hline
        \multirow{2}{*}{$N$}
        &\multicolumn{2}{c|}{RGB} & \multicolumn{2}{c|}{Flow} & \multicolumn{2}{c}{Audio}\\
        \cline{2-7}
        & w/o & w/ & w/o & w/ & w/o & w/\\
        \hline
        1  & 73.1  & 73.4  & 63.7  & 65.2  & 21.3  & 21.5 \\
        2  & 73.6  & 73.9  & 64.6  & 66.1  & 22.3  & 22.3 \\
        4  & 73.8  & 74.2  & 65.0  & 66.5  & 22.8  & 22.9 \\
        8  & \textbf{74.1}  & 74.4  & 65.5  & 66.8  & \textbf{23.3}  & 23.4 \\
        16  & 74.0  & 74.6  & \textbf{65.9}  & 67.1  & \textbf{23.3}  & 23.7 \\
        32  & 73.6  & 74.7  & 65.5  & 67.4  & 23.1  & 23.9 \\
        64  & 73.2  & 74.9  & 65.1  & \textbf{67.5}  & 22.7  & 24.1 \\
        128  & 73.2  & \textbf{75.0}  & 65.1  & \textbf{67.5}  & 22.5  & \textbf{24.2} \\
        \hline
        \end{tabular}
        }
        \caption{Top-1 accuracy (\%) on Kinetics to show the effect of different cluster sizes and training with (w/) or without (w/o) shifting operation for RGB, flow, and audio.}
        \label{tab:kinetics}
    \end{table}

    \subsection{Local Feature Extraction}
        Considering that video is inherently multimodal, we extract three kinds of local features -- RGB, flow, and audio -- to represent the video. We rely on CNNs to extract these features.

        RGB and flow features are extracted from RGB video frames or optical flow images, which are created by stacking the two channels for the horizontal and vertical vector fields \cite{Simonyan2014Two}. For UCF101 and HMDB51, we initialize ResNet-152 \cite{resnet} with a pre-trained ImageNet model and fine-tune it using the frames from training videos and then apply it to extract RGB and flow features. For Kinetics, we rely on Inception-ResNet-v2 \cite{inception-resnet} to extract these features. The RGB model is initialized with a pre-trained ImageNet model and fine-tuned using the training split based on the temporal segment network framework \cite{TSN} with 7 segments. Then the flow model is initialized by the RGB model and also fine-tuned the same way. After training, we can extract local RGB and flow features for every frame.

        To extract audio features, we generate audio spectrogram patches first. For every 10ms, we decompose the signal with a short-time Fourier transform and then rely on aggregated, logarithm-transformed 64 mel-spaced frequency bins following \cite{audiocnn}. Each 96 consecutive bins yield one log-mel ${96\times64}$ spectrogram patch, which can be processed just like an image.
        After this, we can extract audio features using VGG-16 \cite{vgg16} on Kinetics just as for RGB and flow features. For videos without audio, we feed in the average of audio features over the training set.

    \subsection{Local Feature Set Augmentation} \label{sec:augmentation}
        Data augmentation plays a very important role in training neural networks, making use of properties of the data to effectively reduce overfitting. Here, we can similarly exploit the properties of the local feature sets to design new data enhancement methods. Local feature sets are approximately unordered, and most of the time, we do not need to understand the video using all of the local features, since we only need a few key frames to understand video. Hence, when we train the model, we can randomly sample a part of the features from the local feature set, but use all the features during testing. This data augmentation method can reduce the amount of computation during training, effectively prevent overfitting, and allow us to make use of all information during testing.

    \begin{table}[t]
        \centering
        \resizebox{0.36\textwidth}{!}{
        \begin{tabular}{c|c|c|c|c}
        \hline
        $N$ RGB& $N$ Flow& $N$ Audio & Top-1 (\%) &Top-5 (\%) \\
        \hline
        1 & 1 & 1 & 77.9  & 93.6 \\
        4 & 4 & 4 & 78.7  & 94.0 \\
        16 & 16 & 16 & 79.1  & 94.0  \\
        32 & 16 & 16 & 79.2  & 94.0  \\
        32 & 32 & 32 & 79.3  & 93.9  \\
        \textbf{64} & \textbf{32} & \textbf{32} & \textbf{79.4}  & \textbf{94.0}  \\
        64 & 64 & 64 & 79.3  & 94.0  \\
        128 & 128 & 128 & 79.3  & 93.9  \\
        \hline
        \end{tabular}
        }
        \caption{Top-1 and top-5 accuracy (\%) of multimodal integration of different cluster sizes for different modality on Kinetics.}
        \label{tab:kinetics-multimodal}
    \end{table}

    \subsection{Implementation Details}
        In order to reduce overfitting, we apply dropout with probability 0.9 before the final fully connected layer. For local feature set augmentation, we sample 15/15/20 local features during training on UCF101/HMDB51/Kinetics, respectively, and we extract 20/20/25 local features, respectively, at equal intervals during testing. To balance the dataset, we set the sample weight to $1/S$ if a given sample belongs to a class that contains $S$ samples during training. We rely on the RMSPROP algorithm  \cite{Tieleman2012rmsprop} to update parameters with a learning rate of $0.001$ and clip the gradient $\ell_2$-norm of all the parameters to 5 for better convergence.

        \begin{table}[t]
            \centering
            \resizebox{0.38\textwidth}{!}{
            \begin{tabular}{c|c|c|c}
            \hline
            \multicolumn{2}{c|}{Method} & Top-1(\%) & Top-5(\%) \\
            \hline
            \multicolumn{2}{c|}{C3D \cite{C3D}} & 55.6 &79.1\\
            \multicolumn{2}{c|}{3D ResNet \cite{3dresnet}} & 58.0 &81.3\\
            \multicolumn{2}{c|}{Two-Stream I3D* \cite{2si3d}} & 74.2 & 91.3\\
            \multicolumn{2}{c|}{RGB+Flow TSN Inception V3 \cite{TSN}} & 76.6 & 92.4\\
            \hline
            \hline

            \multirow{5}{*}{ \tabincell{l}{RGB}}
            &TSN\cite{TSN} & 73.0   & 90.9\\
            &TS-LSTM (5 seg) \cite{tslstm} & 73.2 & 90.9 \\
            &Temporal-Inception \cite{tslstm} & 73.5 &91.2\\
            &Bi-directional LSTM & 74.0 &91.6 \\
            \cline{2-4}
            &\textbf{Attention Cluster}  & \textbf{75.0}   & \textbf{91.9} \\
            \hline

            \multirow{5}{*}{ \tabincell{l}{Flow}}
            &TSN\cite{TSN} & 66.0    & 86.9 \\
            &TS-LSTM (5 seg) \cite{tslstm} & 65.3 & 86.2 \\
            &Temporal-Inception \cite{tslstm} & 65.4 &86.2\\
            &Bi-directional LSTM & 66.4 &86.9 \\
            \cline{2-4}
            &\textbf{Attention Cluster}  & \textbf{67.5}    &\textbf{87.3} \\
            \hline

            \multirow{5}{*}{ \tabincell{l}{Audio}}
            &TSN\cite{TSN} &21.6    &39.4 \\
            &TS-LSTM (5 seg) \cite{tslstm} & 22.6 & 40.6 \\
            &Temporal-Inception \cite{tslstm} & 22.7 &40.7\\
            &Bi-directional LSTM & 23.4 &41.3 \\
            \cline{2-4}
            &\textbf{Attention Cluster}  &\textbf{24.2}    &\textbf{42.2} \\
            \hline
            \hline

            \multirow{6}{*}{ \tabincell{l}{RGB  \\ + Flow \\ + Audio}}
            &Average & 76.5 &92.7 \\
            &Flatten & 76.7 &92.7 \\
            &TS-LSTM (5 seg) \cite{tslstm} & 77.3 & 93.0 \\
            &Temporal-Inception \cite{tslstm} & 77.6 &93.3\\
            &Bi-directional LSTM & 78.2 &93.5 \\
            \cline{2-4}
            &\textbf{Attention Cluster} &\textbf{79.4} &\textbf{94.0} \\
            \hline
            \end{tabular}
            }
            \caption{Kinetics top-1 and top-5 accuracy (\%) on the validation set, except for results marked with `$\ast$', which were reported based on the test set.}
            \label{tab:results-kinetics}
        \end{table}

    \subsection{Result of Single Modality}

    We explore how many attention units we need to use for a single modality and whether we should use the shifting operation or not to achieve the best results on Kinetics. Based on the previous experiment, we use the weighting function FC1 as default. Table \ref{tab:kinetics} describes the relationship between the Top-1 accuracy of RGB, flow, and audio as single modalities, for different attention cluster sizes $N$, and with or without the shifting operation.

    We find that often with increasing $N$, the accuracy increases first and then decreases without shifting, or remains unchanged with the shifting operation. When $N$ is small, due to limitations of the expressive power of attention models, the accuracy increases with increasing $N$ both with and without the shifting operation. When $N$ is sufficiently large, the expressive power is adequate. Without the shifting operation, for increases in $N$, the number of parameters also increases, and harmful overfitting becomes a serious issue, while the training is also more difficult. This leads to a decrease in accuracy. With the shifting operation, the training remains stable and reliable even for large cluster sizes.

    We observe that the accuracy while using the shifting operation is universally better than without it. This suggests that the shifting operation can  increase the diversity of the attention mechanism effectively, to improve the accuracy. We have also observed that attention clusters with shifting operation converged more quickly than without, similar to the observations on Flash--MNIST.

    As shown in Table \ref{tab:results-kinetics}, comparing to the pretrained TSN models used for feature extraction, our attention clusters achieve excellent improvements of 2.0\% for RGB, 1.5\% for flow, and 2.6\% for audio, in terms of top-1 accuracy. We also find that our results can beat other fusion methods using the same local features.

    \subsection{Result of Multimodal Integration}

    We investigate the effects of various combinations of different attention cluster sizes for multimodal integration. The results are shown in Table \ref{tab:kinetics-multimodal}. We find that we can use smaller cluster sizes for multimodal integration rather than for a single modality. We can achieve the best top-1 accuracy (79.4\%) and the best top-5 accuracy (94.0\%) with 64 attention units for RGB, and 32 for flow and audio.
    We also implement a series of three stream fusion methods using the same local features (see Section \ref{sec:detail-multi} for details). As shown in Table \ref{tab:results-kinetics}, our approach improved over them by a large margin.

         \begin{table}[t]
            \centering
            \resizebox{0.40\textwidth}{!}{
            \begin{tabular}{c|c|c|c}
            \hline
            \multicolumn{2}{c|}{Method} & UCF101(\%) & HMDB51 (\%) \\
            \hline
            \multicolumn{2}{c|}{iDT + FV \cite{iDT}} &85.9  &57.2 \\
            \multicolumn{2}{c|}{iDT + HSV \cite{Peng2016Bag}} &87.9   &61.1 \\
            \hline
            \multicolumn{2}{c|}{EMV-CNN \cite{Zhang2016Real}} &86.4   &- \\
            \multicolumn{2}{c|}{VideoLSTM\cite{Li2016VideoLSTM}} &89.2   &- \\
            \multicolumn{2}{c|}{FSTCN \cite{fstcn}} &88.1   &59.1 \\
            \multicolumn{2}{c|}{TDD+FV \cite{tdd}} &90.3   &63.2 \\
            \multicolumn{2}{c|}{TSN (2 modalities) \cite{TSN}} &94.0    &68.5 \\
            \multicolumn{2}{c|}{Two Stream \cite{Simonyan2014Two}} &88.0   &59.4 \\
            \multicolumn{2}{c|}{Temporal-Inception \cite{tslstm}} & 93.9 &67.5\\
            \multicolumn{2}{c|}{TS-LSTM \cite{tslstm}} & 94.1 & 69.0 \\
            \multicolumn{2}{c|}{Fusion \cite{Feichtenhofer2016Convolutional}} &92.5    &65.4 \\
            \multicolumn{2}{c|}{ST-ResNet \cite{Feichtenhofer2016Spatiotemporal}} & 93.4   &66.4 \\
            \multicolumn{2}{c|}{ActionVLAD \cite{ActionVLAD}} &92.7    &66.9 \\
            \hline
            \multicolumn{2}{c|}{\textbf{Attention Cluster RGB+Flow}} &\textbf{94.6}    &\textbf{69.2} \\
            \hline
            \end{tabular}
            }
            \caption{Mean classification accuracy (\%) comparing with State-of-the-Art methods on UCF101 and HMDB51.}
            \label{tab:results-ucf101}
        \end{table}

    \subsection{Comparison with State-of-the-Art}
        Finally, we compare our method against the state-of-the-art methods.

        On UCF101 and HMDB, our approach obtains robust improvements over the two-stream fusion results for CNNs.
        As shown in Table \ref{tab:results-ucf101}, our approach can achieve competitive results in comparison with existing published methods \cite{iDT,Peng2016Bag,Zhang2016Real,Simonyan2014Two,fstcn,Li2016VideoLSTM,tdd,Feichtenhofer2016Convolutional,C3D,TSN,Feichtenhofer2016Spatiotemporal,ActionVLAD,tslstm}.

        On Kinetics, as shown in Table \ref{tab:results-kinetics}, we compare our method against many published results \cite{C3D,3dresnet,2si3d,TSN}. Since the local feature extractors are trained using TSN \cite{TSN}, the results with CNNs are already very strong.
        The implemented three stream fusion methods also act as a strong baseline.
        Our approach again enjoys great improvements over all of them and obtains the start-of-the-art result.

\section{Conclusion}
    To explore the potential of pure attention networks for video classification, a new architecture based on attention clusters with a shifting operation is proposed to integrate local feature sets.
    We analyze and visualize attention on the proposed Flash--MNIST to get a better understand of how our attention clusters work. We also have conducted experiments on three well-known video classification datasets and find that this architecture can achieve excellent results for a single modality or integrating multiple modalities, while also accelerating the training phase.

    In terms of future work, we hope to apply this architecture to low-level local features and assess to what extent it can uncover relationships between features in different spatial coordinates. We further hope to integrate it into end-to-end-trained networks.

\clearpage
\section*{Appendices}

\appendix
\renewcommand{\appendixname}{Appendix~\Alph{section}}
\section{Details of the Flash--MNIST Experiments} \label{sec:detail-mnist}
    In this section, we provide more details of the dataset generation, local feature extraction, and implementation of our attention cluster training on Flash--MNIST.

    \subsection{Dataset Generation}
        We generate the 102,400 training samples and 10,240 test samples that constitute our proposed Flash--MNIST dataset as follows:

        \begin{enumerate}
        \item \textbf{Generating frames with noisy background}: We randomly generate 25 different $28 \times 28$ noise frames for each video sample. First, since the each pixel of MNIST images is represented by an integer value in [0,255], we compute the distribution over such pixel intensity values in the MNIST training data, i.e., we count the frequency of each such integer in the training set. Then, for each pixel in each frame to be generated for our Flash--MNIST data, we randomly sample an integer value in $[0,255]$ in accordance with the computed probability distribution.

        \item \textbf{Randomly sampling digits}: We randomly sample a category from the set of 1024 possible categories in Flash--MNIST. Then, for each digit belonging to the sampled category, we randomly sample one or two corresponding images from MNIST, either from the training or the testing set, depending on target video. Thus, we may sample $0-20$ MNIST digit images per target video.

        \item \textbf{Inserting digits into random frames}: We randomly pick the frame into which each sampled image shall be inserted. Then, we overlay the digit images on the random background by keeping the maximum value for each pixel.

        \end{enumerate}

    \subsection{Local Feature Extraction}
        To extract local features, we apply the following steps:
        \begin{enumerate}
        \item \textbf{Collecting samples for pretraining}: For each sample in the MNIST training split, we randomly generate 5 noise background images and place a digit on them. We also generate 30,000 noisy backgrounds without any digit. The goal is to classify each sample to one of 11 categories, digits in $\{0,\dots,9\}$ or just a noisy background.

        \item \textbf{Training the network for local feature extraction}: We apply CNNs for frame classification. The CNNs consist of 2 successive convolutional layers with $5 \times 5$ kernels, 10/20 filters followed by \emph{relu} activations and max-pooling with stride 2, one fully connected layer with 50 hidden units followed \emph{relu} activations, and finally a fully connected layer with 11 hidden units followed by \emph{softmax}. We use Adam optimization \cite{Kingma2014Adam} with a learning rate of $0.001$ to update the parameters and train for 10 epochs on the collected samples.

        \item \textbf{Extracting local features for frames}: We apply the pretrained CNNs except for the last fully connected layer to extract local features with a dimensionality of 50 for frames in Flash--MNIST.
        \end{enumerate}

    \subsection{Attention Cluster Training Details}
        After extracting local features for each sample, we apply our attention clusters approach to obtain the global feature representation and then use a fully connected layer with $1024$ hidden units followed by a \emph{softmax} layer for classification. In order to reduce overfitting, we apply dropout with probability 0.5 before the final fully connected layer.
        We rely on the Adam algorithm to update parameters with a learning rate of $0.001$ and at most train for 100 epochs.

\section{Details of Multimodal Integration Methods} \label{sec:detail-multi}
    In this section, we introduce the network structure for the comparison of multimodal integration methods described in Section 5.6. The Average and Flatten methods serve as basic baselines. TS-LSTM \cite{tslstm}, Temporal-Inception \cite{tslstm}, and Bi-directional LSTMs provide baselines with temporal modeling.

     \begin{itemize}
        \item \textbf{Average}: The Average baseline is the most straightforward method. This method generates global features by concatenating the global averages of each modality. The global feature is used for classification through a fully connected layer.
        \item \textbf{Flatten}: The Flatten method is also a straightforward one. This method first flattens the local features of each modality. This means that we can obtain an $LM$ length output for a local feature set of dimensionality $L \times M$. Then, the outputs of three modalities are concatenated for classification.
        \item \textbf{TS-LSTM and Temporal-Inception}: We simply extend the two-modality integration methods described in the original work \cite{tslstm} to a three-modality version. We set the number of segments to 5 for TS-LSTM.
        \item \textbf{Bi-directional LSTMs}: For local features of the RGB/flow/audio signals, we apply a batch normalization layer \cite{bn} followed by bi-directional LSTMs \cite{lstm} with 1024/512/512 hidden units, respectively. The averages of the output hidden states of each bi-directional LSTM are then concatenated for classification.
     \end{itemize}


\begin{thebibliography}{10}\itemsep=-1pt

\bibitem{youtube8m}
S.~{Abu-El-Haija}, N.~{Kothari}, J.~{Lee}, P.~{Natsev}, G.~{Toderici},
  B.~{Varadarajan}, and S.~{Vijayanarasimhan}.
\newblock {YouTube-8M: A Large-Scale Video Classification Benchmark}.
\newblock {\em ArXiv e-prints}, Sept. 2016.

\bibitem{ba2014multiple}
J.~Ba, V.~Mnih, and K.~Kavukcuoglu.
\newblock Multiple object recognition with visual attention.
\newblock {\em arXiv preprint arXiv:1412.7755}, 2014.

\bibitem{bahdanau2014neural}
D.~Bahdanau, K.~Cho, and Y.~Bengio.
\newblock Neural machine translation by jointly learning to align and
  translate.
\newblock {\em arXiv preprint arXiv:1409.0473}, 2014.

\bibitem{bian2017revisiting}
Y.~Bian, C.~Gan, X.~Liu, F.~Li, X.~Long, Y.~Li, H.~Qi, J.~Zhou, S.~Wen, and
  Y.~Lin.
\newblock Revisiting the effectiveness of off-the-shelf temporal modeling
  approaches for large-scale video classification.
\newblock {\em arXiv preprint arXiv:1708.03805}, 2017.

\bibitem{2si3d}
J.~{Carreira} and A.~{Zisserman}.
\newblock {Quo Vadis, Action Recognition? A New Model and the Kinetics
  Dataset}.
\newblock {\em ArXiv e-prints}, May 2017.

\bibitem{Cheng2016Long}
J.~{Cheng}, L.~{Dong}, and M.~{Lapata}.
\newblock {Long Short-Term Memory-Networks for Machine Reading}.
\newblock {\em ArXiv e-prints}, Jan. 2016.

\bibitem{donahue2015long}
J.~Donahue, L.~Anne~Hendricks, S.~Guadarrama, M.~Rohrbach, S.~Venugopalan,
  K.~Saenko, and T.~Darrell.
\newblock Long-term recurrent convolutional networks for visual recognition and
  description.
\newblock In {\em CVPR}, pages 2625--2634, 2015.

\bibitem{Feichtenhofer2016Spatiotemporal}
C.~Feichtenhofer, A.~Pinz, and R.~P. Wildes.
\newblock Spatiotemporal residual networks for video action recognition.
\newblock In {\em NIPS}, 2016.

\bibitem{Feichtenhofer2016Convolutional}
C.~Feichtenhofer, A.~Pinz, and A.~Zisserman.
\newblock Convolutional two-stream network fusion for video action recognition.
\newblock In {\em CVPR}, 2016.

\bibitem{gan2016webly}
C.~Gan, C.~Sun, L.~Duan, and B.~Gong.
\newblock Webly-supervised video recognition by mutually voting for relevant
  web images and web video frames.
\newblock In {\em ECCV}, pages 849--866, 2016.

\bibitem{gan2015devnet}
C.~Gan, N.~Wang, Y.~Yang, D.-Y. Yeung, and A.~G. Hauptmann.
\newblock {DevNet}: A deep event network for multimedia event detection and
  evidence recounting.
\newblock In {\em CVPR}, pages 2568--2577, 2015.

\bibitem{gan2016you}
C.~Gan, T.~Yao, K.~Yang, Y.~Yang, and T.~Mei.
\newblock You lead, we exceed: Labor-free video concept learning by jointly
  exploiting web videos and images.
\newblock In {\em CVPR}, pages 923--932, 2016.

\bibitem{ActionVLAD}
R.~Girdhar, D.~Ramanan, A.~Gupta, J.~Sivic, and B.~Russell.
\newblock Actionvlad: Learning spatio-temporal aggregation for action
  classification.
\newblock In {\em CVPR}, 2017.

\bibitem{3dresnet}
K.~{Hara}, H.~{Kataoka}, and Y.~{Satoh}.
\newblock {Learning Spatio-Temporal Features with 3D Residual Networks for
  Action Recognition}.
\newblock {\em ArXiv e-prints}, Aug. 2017.

\bibitem{resnet}
K.~He, X.~Zhang, S.~Ren, and J.~Sun.
\newblock Deep residual learning for image recognition.
\newblock In {\em CVPR}, 2016.

\bibitem{audiocnn}
S.~{Hershey}, S.~{Chaudhuri}, D.~P.~W. {Ellis}, J.~F. {Gemmeke}, A.~{Jansen},
  R.~{Channing Moore}, M.~{Plakal}, D.~{Platt}, R.~A. {Saurous}, B.~{Seybold},
  M.~{Slaney}, R.~J. {Weiss}, and K.~{Wilson}.
\newblock {CNN Architectures for Large-Scale Audio Classification}.
\newblock {\em ArXiv e-prints}, Sept. 2016.

\bibitem{lstm}
S.~Hochreiter and J.~Schmidhuber.
\newblock Long short-term memory.
\newblock {\em Neural Computation}, 9(8):1735--1780, 1997.

\bibitem{bn}
S.~{Ioffe} and C.~{Szegedy}.
\newblock {Batch Normalization: Accelerating Deep Network Training by Reducing
  Internal Covariate Shift}.
\newblock {\em ArXiv e-prints}, Feb. 2015.

\bibitem{Karpathy2014Large}
A.~Karpathy, G.~Toderici, S.~Shetty, T.~Leung, R.~Sukthankar, and F.~F. Li.
\newblock Large-scale video classification with convolutional neural networks.
\newblock In {\em CVPR}, pages 1725--1732, 2014.

\bibitem{Kingma2014Adam}
D.~P. Kingma and J.~Ba.
\newblock Adam: A method for stochastic optimization.
\newblock In {\em ICLR}, 2015.

\bibitem{alexnet}
A.~Krizhevsky, I.~Sutskever, and G.~E. Hinton.
\newblock Imagenet classification with deep convolutional neural networks.
\newblock In {\em NIPS}, pages 1097--1105, 2012.

\bibitem{Kuehne2011HMDB}
H.~Kuehne, H.~Jhuang, E.~Garrote, T.~Poggio, and T.~Serre.
\newblock Hmdb: A large video database for human motion recognition.
\newblock In {\em ICCV}, pages 2556--2563, 2011.

\bibitem{Li2016VideoLSTM}
Z.~{Li}, E.~{Gavves}, M.~{Jain}, and C.~G.~M. {Snoek}.
\newblock {VideoLSTM Convolves, Attends and Flows for Action Recognition}.
\newblock {\em ArXiv e-prints}, July 2016.

\bibitem{Lin2017A}
Z.~{Lin}, M.~{Feng}, C.~{Nogueira dos Santos}, M.~{Yu}, B.~{Xiang}, B.~{Zhou},
  and Y.~{Bengio}.
\newblock {A Structured Self-attentive Sentence Embedding}.
\newblock {\em ArXiv e-prints}, Mar. 2017.

\bibitem{tslstm}
C.-Y. {Ma}, M.-H. {Chen}, Z.~{Kira}, and G.~{AlRegib}.
\newblock {TS-LSTM and Temporal-Inception: Exploiting Spatiotemporal Dynamics
  for Activity Recognition}.
\newblock {\em ArXiv e-prints}, Mar. 2017.

\bibitem{mnih2014recurrent}
V.~Mnih, N.~Heess, A.~Graves, et~al.
\newblock Recurrent models of visual attention.
\newblock In {\em Advances in neural information processing systems}, pages
  2204--2212, 2014.

\bibitem{Ng2015Beyond}
Y.~H. Ng, M.~Hausknecht, S.~Vijayanarasimhan, O.~Vinyals, R.~Monga, and
  G.~Toderici.
\newblock Beyond short snippets: Deep networks for video classification.
\newblock In {\em CVPR}, 2015.

\bibitem{2017arXiv170504304P}
R.~{Paulus}, C.~{Xiong}, and R.~{Socher}.
\newblock {A Deep Reinforced Model for Abstractive Summarization}.
\newblock {\em ArXiv e-prints}, May 2017.

\bibitem{Peng2016Bag}
X.~Peng, L.~Wang, X.~Wang, and Y.~Qiao.
\newblock Bag of visual words and fusion methods for action recognition:
  Comprehensive study and good practice.
\newblock {\em Computer Vision and Image Understanding}, 150(C):109--125, 2016.

\bibitem{sharma2015action}
S.~{Sharma}, R.~{Kiros}, and R.~{Salakhutdinov}.
\newblock {Action Recognition using Visual Attention}.
\newblock {\em ArXiv e-prints}, Nov. 2015.

\bibitem{Simonyan2014Two}
K.~Simonyan and A.~Zisserman.
\newblock Two-stream convolutional networks for action recognition in videos.
\newblock {\em NIPS}, 2014.

\bibitem{vgg16}
K.~{Simonyan} and A.~{Zisserman}.
\newblock {Very Deep Convolutional Networks for Large-Scale Image Recognition}.
\newblock {\em ArXiv e-prints}, Sept. 2014.

\bibitem{ucf101}
K.~{Soomro}, A.~{Roshan Zamir}, and M.~{Shah}.
\newblock {UCF101: A Dataset of 101 Human Actions Classes From Videos in The
  Wild}.
\newblock {\em ArXiv e-prints}, Dec. 2012.

\bibitem{Srivastava2015Unsupervised}
N.~Srivastava, E.~Mansimov, and R.~Salakhutdinov.
\newblock Unsupervised learning of video representations using lstms.
\newblock In {\em ICML}, 2015.

\bibitem{fstcn}
L.~Sun, K.~Jia, D.~Y. Yeung, and B.~E. Shi.
\newblock Human action recognition using factorized spatio-temporal
  convolutional networks.
\newblock In {\em ICCV}, 2015.

\bibitem{inception-resnet}
C.~{Szegedy}, S.~{Ioffe}, V.~{Vanhoucke}, and A.~{Alemi}.
\newblock {Inception-v4, Inception-ResNet and the Impact of Residual
  Connections on Learning}.
\newblock {\em ArXiv e-prints}, Feb. 2016.

\bibitem{inception}
C.~Szegedy, W.~Liu, Y.~Jia, P.~Sermanet, S.~Reed, D.~Anguelov, D.~Erhan,
  V.~Vanhoucke, and A.~Rabinovich.
\newblock Going deeper with convolutions.
\newblock In {\em CVPR}, pages 1--9, 2015.

\bibitem{Tieleman2012rmsprop}
T.~Tieleman and G.~Hinton.
\newblock Lecture 6.5-rmsprop: Divide the gradient by a running average of its
  recent magnitude.
\newblock {\em COURSERA: Neural Networks for Machine Learning}, 2012.

\bibitem{C3D}
D.~Tran, L.~Bourdev, R.~Fergus, L.~Torresani, and M.~Paluri.
\newblock Learning spatiotemporal features with 3d convolutional networks.
\newblock In {\em ICCV}, 2015.

\bibitem{Varol2017Long}
G.~Varol, I.~Laptev, and C.~Schmid.
\newblock Long-term temporal convolutions for action recognition.
\newblock {\em IEEE Transactions on Pattern Analysis and Machine Intelligence},
  PP(99):1--1, 2017.

\bibitem{iDT}
H.~Wang and C.~Schmid.
\newblock Action recognition with improved trajectories.
\newblock In {\em ICCV}, 2013.

\bibitem{tdd}
L.~Wang, Y.~Qiao, and X.~Tang.
\newblock Action recognition with trajectory-pooled deep-convolutional
  descriptors.
\newblock In {\em CVPR}, pages 4305--4314, 2015.

\bibitem{TSN}
L.~Wang, Y.~Xiong, Z.~Wang, Y.~Qiao, D.~Lin, X.~Tang, and L.~V. Gool.
\newblock Temporal segment networks: Towards good practices for deep action
  recognition.
\newblock In {\em ECCV}, 2016.

\bibitem{optical-flow}
C.~Zach, T.~Pock, and H.~Bischof.
\newblock A duality based approach for realtime tv-l1 optical flow.
\newblock {\em Annunal Symp. German Association Pattern Recognition},
  4713(5):214--223, 2007.

\bibitem{Zhang2016Real}
B.~Zhang, L.~Wang, Z.~Wang, Y.~Qiao, and H.~Wang.
\newblock Real-time action recognition with enhanced motion vector cnns.
\newblock In {\em CVPR}, 2016.

\end{thebibliography}
\end{document}